\documentclass[prologue,dvipsnames,sigconf,nonacm]{acmart}
\usepackage{color,soul}
\usepackage{subfig}
\usepackage{todonotes}
\usepackage{bbm}
\usepackage{enumitem}
\graphicspath{ {./img/} }
\DeclareRobustCommand{\hlcyan}[1]{{\sethlcolor{lime}\hl{#1}}}
\DeclareRobustCommand{\hlblue}[1]{{\sethlcolor{Lavender}\hl{#1}}}
\definecolor{aqua}{rgb}{0.0, 1.0, 1.0}
\DeclareRobustCommand{\hlpurple}[1]{{\sethlcolor{aqua}\hl{#1}}}

\AtBeginDocument{%
  \providecommand\BibTeX{{%
    \normalfont B\kern-0.5em{\scshape i\kern-0.25em b}\kern-0.8em\TeX}}}

\setcopyright{acmcopyright}
\copyrightyear{2022}
\acmYear{2022}
\acmDOI{XXXXXXX.XXXXXXX}
\begin{document}

\title{Cross-Domain Multi-Task Learning for Sequential Sentence Classification in Research Papers}

\author{Arthur Brack}
\orcid{0000-0002-1428-5348}
\affiliation{%
  \institution{TIB -- Leibniz Information Centre for Science and Technology \& Leibniz University}
  \city{Hannover}
  \country{Germany}}
\email{Arthur.Brack@tib.eu}

\author{Anett Hoppe}
\orcid{0000-0002-1452-9509}
\affiliation{%
  \institution{TIB -- Leibniz Information Centre for Science and Technology \& L3S Research Center}
  \city{Hannover}
  \country{Germany}}
\email{Anett.Hoppe@tib.eu}

\author{Pascal Buscherm\"ohle}
\affiliation{%
  \institution{Leibniz University}
  \city{Hannover}
  \country{Germany}}

\author{Ralph Ewerth}
\orcid{0000-0003-0918-6297}
\affiliation{%
  \institution{TIB -- Leibniz Information Centre for Science and Technology \& L3S Research Center}
  \city{Hannover}
  \country{Germany}}
\email{Ralph.Ewerth@tib.eu}

\renewcommand{\shortauthors}{Brack et al.}


\begin{abstract}
Sequential sentence classification deals with the categorisation of sentences based on their content and context. 
Applied to scientific texts, it enables the automatic structuring of research papers and the improvement of academic search engines.
However, previous work has not investigated the potential of transfer learning for sentence classification across different scientific domains and the issue of different text structure of full papers and abstracts. 
In this paper, we derive seven related research questions and present several contributions to address them: 
First, we suggest a novel uniform deep learning architecture and multi-task learning for cross-domain sequential sentence classification in scientific texts. 
Second, we tailor two common transfer learning methods, sequential transfer learning and multi-task learning, to deal with the challenges of the given task. 
Semantic relatedness of tasks is a prerequisite for successful transfer learning of neural models.
Consequently, our third contribution is an approach to semi-automatically identify semantically related classes from different annotation schemes and we present an analysis of four annotation schemes.
Comprehensive experimental results indicate that models, which are trained on datasets from different scientific domains, benefit from one another when using the proposed multi-task learning architecture. 
We also report comparisons with several state-of-the-art approaches.
Our approach outperforms the state of the art on full paper datasets significantly while being on par for datasets consisting of abstracts.
\end{abstract}




\maketitle

\section{Introduction}

To search relevant research papers for a particular field is a core activity of researchers.
Scientists usually use academic search engines and skim through the text of the found articles to assess their relevance.
However, academic search engines cannot assist researchers adequately in these tasks since most research papers are plain PDF files and not machine-interpretable \citep{Brack2021IJDL,Safder2019,Xiong2017ExplicitSR}. 
The exploding number of published articles aggravates this situation further~\citep{bornmann15growth}. 
Therefore, automatic approaches to structure research papers are highly desired.

\begin{figure}[t]
\vspace{2em}
\footnotesize
\centering%
\begin{tabular}{| p{\dimexpr0.47\textwidth-2\tabcolsep-\arrayrulewidth\relax}|   }
\hline
\\
\hlcyan{Gamification has the potential to improve the quality of learning by better engaging students with learning activities.} \hl{Our objective in this study is to evaluate a gamified learning activity along the dimensions of learning, engagement, and enjoyment.} \hlblue{The activity made use of a gamified multiple choice quiz implemented as a software tool and was trialled in three undergraduate IT-related courses. A questionnaire survey was used to collect data to gauge levels of learning, engagement, and enjoyment.} \hlpurple{Results show that there was some degree of engagement and enjoyment. The majority of participants (77.63 per cent) reported that they were engaged enough to want to complete the quiz and 46.05 per cent stated they were happy while playing the quiz...}\\
\\
\hline
\end{tabular}
\caption{An annotated abstract taken from the CSABSTRUCT dataset \cite{Cohan2019PretrainedLM}, in which sentences describing the \emph{background} (green), \emph{objectives} (yellow), \emph{methods} (magenta), and \emph{results} (cyan) of the paper are coloured.}
\label{fig:highlight-example}
\end{figure}

Sequential sentence classification targets the categorisation of sentences by their semantic content or function. In research papers, this can be used to classify sentences by their contribution to the article's content, e.g. to determine if a certain sentence contains information about the research work's objective, methods or results \citep{Dernoncourt2017PubMed2R}.
Figure \ref{fig:highlight-example} shows an example of an abstract with classified sentences.
Such a semantification of sentences can help to focus on relevant elements of text and thus assist information retrieval systems \citep{Neves2019EvaluationOS,Safder2019} or knowledge graph population~\citep{Oelen21}.
The task is called \emph{sequential} to distinguish it from the general \emph{sentence classification} task where a sentence is classified in isolation, i.e.\ without using local context. 
However, in research papers the meaning of a sentence is often informed by the context from neighbouring sentences, 
e.g.\ sentences describing the methods usually precede sentences about results.

Several approaches have been proposed for \emph{sequential sentence classification} (e.g.\ \citep{Asadi2019AutomaticZI,Jin2018HierarchicalNN,Shang21}), and several datasets were annotated for various scientific domains (e.g.\ \citep{Dernoncourt2017PubMed2R,Fisas2015OnTD,Goncalves20,SteadSBV19}). The datasets contain either abstracts or full papers and were annotated with domain-specific sentence classes. 
However, research infrastructures usually support multiple scientific domains. Therefore, stakeholders of digital libraries are interested in a uniform solution that enables the combination of these datasets to improve the overall accuracy. For this purpose, this paper explores the following research questions.

First, although some approaches propose transfer learning for the scientific domain~\cite{Banerjee20,Brack2021Coref,Gupta21,park-caragea-2020-scientific}, the field lacks a comprehensive empirical study on transfer learning across different scientific domains for \emph{sequential sentence classification}.
Transfer learning enables the combination of knowledge from multiple datasets to improve classification performance and thus to reduce annotation costs.
The annotation of scientific text is particularly costly since it demands expertise in the article’s domain~\citep{augenstein2017semeval,Brack2020DomainindependentEO,gabor2018semeval}.
However, studies revealed that the success of transferring neural models depends largely on the relatedness of the tasks, and transfer learning with unrelated tasks may even degrade the performance \citep{Mou2016HowTA,Ruder2019Neural,Pan10,Semwal2018}.
Two tasks are related if there exists some implicit or explicit relationship between the feature spaces \citep{Pan10}. 
On the other hand, every scientific domain is characterised by its specific terminology and phrasing, which yields different feature spaces. Thus, it is not clear to which extent datasets from different scientific disciplines are related.
This raises the following research questions (RQ) for the task of sequential sentence classification: 
\begin{itemize}[nosep]
\setlength\itemsep{-1em}
\item[RQ1:] To which extent are datasets from different scientific domains semantically related? \\
\item[RQ2:] Which transfer learning approach works best?  \\
\item[RQ3:] Which neural network layers are transferable under which constraints? \\
\item[RQ4:] Is it beneficial to train a multi-task model with multiple datasets? \\
\vspace{-1em}
\end{itemize}

Normally, every dataset has a domain-specific annotation scheme that consists of a set of associated sentence classes. 
This raises the second set of research questions with regard to the consolidation of these annotation schemes. 
Prior work \citep{Liakata2010CorporaFT} annotated a dataset multiple times with different schemes, and analysed the multivariate frequency distributions of the classes. They found that the investigated schemes are complementary and should be combined.
However, annotating datasets multiple times is costly. 
To support the consolidation of different annotation schemes across domains, we examine the following RQs:
\begin{itemize}[nosep]
\setlength\itemsep{-1em}
\item[RQ5:] Can a model trained with multiple datasets recognise the semantic relatedness of classes from different annotation schemes?\\
\item[RQ6:] Can we derive a consolidated, domain-independent annotation scheme and use that scheme to compile a new dataset to train a domain-independent model? \\
\vspace{-1em}
\end{itemize}

Finally, current approaches for sequential sentence classification are designed either for abstracts or full papers. 
One reason is that these text types follow rather different structures: In abstracts, different sentence classes directly follow one another normally. The general paper text, however, exhibits longer passages without change of the semantic sentence class.
Typically, deep learning is used for abstracts \citep{Cohan2019PretrainedLM,Goncalves20,Dernoncourt2016NeuralNF,Jin2018HierarchicalNN,Shang21,yamada-etal-2020-sequential} since presumably more training data are available, whereas for full papers, also called \emph{zone identification}, hand-crafted features and linear models have been suggested \citep{Asadi2019AutomaticZI,Badie2018ZoneIB,Fisas2015OnTD,liakata2012automatic}. 
However, deep learning approaches have also been applied successfully to full papers in related tasks such as argumentation mining \citep{lauscher2018a}, document summarisation~\citep{AbuRaed2020,cohan-etal-2018-discourse,DeYoung21,ghosh-roy-etal-2020-summaformers}, or n-ary relation extraction \citep{Jia2019,friedrich-etal-2020-sofc,Kabongo21}. 
Thus, the potential of deep learning has not been fully exploited yet for sequential sentence classification on full papers, and no unified solution for abstracts as well as full papers exists. This raises the RQ: 
\begin{itemize}[nosep]
\setlength\itemsep{-1em}
\item[RQ7:] Can a unified deep learning approach be applied to text types with very different structures like abstracts or full papers?\\
\vspace{-1em}
\end{itemize}

In this paper, we investigate these research questions and present the following contributions:
(1)~We introduce a novel multi-task learning framework for sequential sentence classification.
(2)~Furthermore, we propose and evaluate an approach to semi-automat\-ically identify semantically related classes from different annotation schemes and present an analysis of four annotation schemes.
Based on the analysis, we suggest a domain-independent annotation scheme and compile a new dataset that enables to classify sentences in a domain-independent manner.
(3)~Our proposed unified deep learning approach can handle both text types, abstracts and full papers, despite their structural differences.
(4)~To facilitate further research, we make our source code publicly available: \url{https://github.com/arthurbra/sequential-sentence-classification}.

Comprehensive experimental results demonstrate that our multi-task learning approach successfully makes use of datasets from different scientific domains, with different annotation schemes, that contain abstracts or full papers.
In particular, we outperform state-of-the-art approaches for full paper datasets significantly, while obtaining competitive results for datasets of abstracts.

The remainder of the paper is organised as follows: Section 2 summarises related work on sentence classification in research papers and transfer learning in NLP. Our proposed approaches are presented in Section 3. The setup and results of our experimental evaluation are reported in Section 4 and 5, while Section 6 concludes the paper and outlines areas of future work.

\section{Related Work}
\label{sec:related_work}
This section outlines datasets for sentence classification in scientific texts and describes machine learning methods for this task. 
Furthermore, we briefly review transfer learning methods. 
For a more comprehensive overview about information extraction from scientific text, we refer to \citet{Brack2021IJDL} and \citet{Nasar2018InformationEF}.

\subsection{Sequential Sentence Classification in Scientific Text}
\label{sec:ml_approaches}

\paragraph{Datasets:}
As depicted in Table~\ref{tab:comparison_sentence_classification}, annotated benchmark datasets for sentence classification in research papers come from various domains, e.g. PubMed-20k~\citep{Dernoncourt2017PubMed2R} consists of biomedical randomised controlled trials, NICTA-PIBOSO~\citep{Kim2011AutomaticCO} comes from evidence-based medicine, Dr. Inventor dataset~\citep{Fisas2015OnTD} from computer graphics, and the ART/Core Scientific Concepts (CoreSC) dataset \citep{Liakata2010CorporaFT} from chemistry and biochemistry. Most datasets cover only abstracts, while ART/CoreSC and Dr. Inventor cover full papers. Furthermore, each dataset has 5 to 11 different sentence classes, which are more domain-independent (e.g. Background, Methods, Results, Conclusions) or more domain-specific (e.g. Intervention, Population \citep{Kim2011AutomaticCO}, or Hypothesis, Model, Experiment \citep{Liakata2010CorporaFT}).

\begin{table*}[t]
\footnotesize
\caption{Characteristics of benchmark datasets for sentence classification in research papers.}
\label{tab:comparison_sentence_classification}
\begin{tabular}{l|l|r|l|l}
\textbf{Dataset}               & \textbf{Domains}                                                                                  & \textbf{\# Papers} & \textbf{Text Type}  & \textbf{Sentence Classes}    \\ \hline
PubMed-20k     \citep{Dernoncourt2017PubMed2R} & Biomedicine                                                                              & 20,000           & abstracts & \begin{tabular}[c]{@{}l@{}}Background, Objective, Methods, Results, Conclusion\end{tabular} \\ \hline
NICTA-PIBOSO   \citep{Kim2011AutomaticCO}      & Biomedicine                                                                              & 1,000            & abstracts & \begin{tabular}[c]{@{}l@{}}Background, Intervention, Study, Population, Outcome, Other\end{tabular} \\ \hline
CSABSTRUCT     \citep{Cohan2019PretrainedLM}   & Computer Science                                                                         & 2,189            & abstracts & \begin{tabular}[c]{@{}l@{}}Background, Objective, Method, Result, Other\end{tabular} \\ \hline
CS-Abstracts   \citep{Goncalves20}             & Computer Science                                                                         & 654              & abstracts & \begin{tabular}[c]{@{}l@{}}Background, Objective, Methods, Results, Conclusions\end{tabular} \\ \hline
Emerald 100k   \citep{SteadSBV19}              & \begin{tabular}[c]{@{}l@{}}Management, Engineering, \\ Information Science\end{tabular}   & 103,457          & abstracts & \begin{tabular}[c]{@{}l@{}}Purpose, Design/methodology/approach, Findings, Originality/value,\\Social implications,Practical implications, Research limitations/implications\end{tabular} \\ \hline
MAZEA          \citep{Dayrell2012RhetoricalMD} & \begin{tabular}[c]{@{}l@{}}Physics, Engineering\\ Life and Health Sciences\end{tabular} & 1,335            & abstracts & \begin{tabular}[c]{@{}l@{}}Background, Gap, Purpose, Method, Result, Conclusion\end{tabular} \\ \hline                              
Dr. Inventor   \citep{Fisas2015OnTD}           & Computer Graphics                                                                        & 40               & full paper & \begin{tabular}[c]{@{}l@{}}Background, Challenge, Approach, Outcome, Future Work\end{tabular} \\ \hline
ART/CoreSC     \citep{Liakata2010CorporaFT}    & \begin{tabular}[c]{@{}l@{}}Chemistry\\ Computational Linguistic\end{tabular}             & 225              & full paper & \begin{tabular}[c]{@{}l@{}}Background, Motivation, Goal, Hypothesis, Object, Model, Method,\\Experiment, Result, Observation, Conclusion\end{tabular} 
\end{tabular}
\end{table*}

\paragraph{Approaches for Abstracts:}
\label{sec:approaches_abstract}

Deep learning has been the preferred approach for sentence classification in  abstracts in recent years~\citep{Cohan2019PretrainedLM,Goncalves20,Dernoncourt2016NeuralNF,Jin2018HierarchicalNN,Shang21,yamada-etal-2020-sequential}. These approaches follow a common \emph{hierarchical sequence labelling architecture}: (1) a word embedding layer encodes tokens of a sentence to word embeddings, (2) a sentence encoder transforms the word embeddings of a sentence to a sentence representation, (3) a context enrichment layer enriches all sentence representations of the abstract with context from surrounding sentences, and (4) an output layer predicts the label sequence. 

As depicted in Table~\ref{tab:comparison_approaches_abstracts}, the approaches vary in different implementations of the layers. The approaches use different kinds of word embeddings, e.g.\ Global Vectors (GloVe) \citep{Pennington2014GloveGV}, Word2Vec~\citep{Mikolov13Word2Vec}, or SciBERT~\citep{Beltagy2019SciBERTPC} that is BERT~\citep{Devlin2018BERTPO} pre-trained on scientific text.
For sentence encoding, a bidirectional long short-term memory (Bi-LSTM)~\citep{Hochreiter1997LongSM} or a convolutional neural network (CNN) with various pooling strategies are utilised, while 
\citet{yamada-etal-2020-sequential} and \citet{Shang21} use the classification token ([CLS]) of BERT or SciBERT. 
To enrich sentences with further context, a recurrent neural network such as a Bi-LSTM or bidirectional gated recurrent unit (Bi-GRU)~\citep{ChoMGBBSB14} is used. \citet{Shang21} additionally exploit an attention-mechanism across sentences; however, it introduces quadratic runtime complexity that depends on the number of sentences.
A conditional random field (CRF) \citep{Lafferty2001ConditionalRF} is mostly used as an output layer to capture the interdependence between classes. \citet{yamada-etal-2020-sequential} form spans of sentence representations and Semi-Markov CRFs to predict the label sequence by considering all possible span sequences of various lengths. 
Thus, their approach can better label longer continuous sentences but is computationally more expensive than a CRF.
\citet{Cohan2019PretrainedLM} obtain contextual sentence representations directly by fine-tuning SciBERT and utilising the separation token ([SEP]) of SciBERT. However, their approach can process only about 10 sentences at once since BERT supports sequences of up to 512 tokens only.

\begin{table}[tb]
\footnotesize
\caption[Comparison of deep learning approaches for sequential sentence classification in abstracts]{Comparison of deep learning approaches for sequential sentence classification in abstracts.}
\label{tab:comparison_approaches_abstracts}
\begin{tabular}{l|l|l|l|l}
\textbf{Approach}                                                       & \begin{tabular}[c]{@{}l@{}}\textbf{Word} \\ \textbf{embedd.}\end{tabular}        & \begin{tabular}[c]{@{}l@{}}\textbf{Sentence} \\  \textbf{encoding}\end{tabular}              & \begin{tabular}[c]{@{}l@{}}\textbf{Context} \\ \textbf{enrichm.}\end{tabular} & \begin{tabular}[c]{@{}l@{}}\textbf{Output} \\ \textbf{layer}\end{tabular}    \\ \hline
\begin{tabular}[c]{@{}l@{}}Dernoncourt \\ and Lee (2016) \cite{Dernoncourt2016NeuralNF} \end{tabular} & \begin{tabular}[c]{@{}l@{}}Char. Emb.\\ + GloVe\end{tabular} & \begin{tabular}[c]{@{}l@{}}Bi-LSTM/\\ concat.\end{tabular}                                                                  & -                                                             & CRF                                                        \\ \hline
\begin{tabular}[c]{@{}l@{}}Jin and Szolovits\\(2018) \cite{Jin2018HierarchicalNN}  \end{tabular}   & \begin{tabular}[c]{@{}l@{}}Bio\\word2vec\end{tabular}                                                      & \begin{tabular}[c]{@{}l@{}}Bi-LSTM/ \\ att. pooling\end{tabular} & Bi-LSTM                                                       & CRF                                                        \\ \hline
\begin{tabular}[c]{@{}l@{}}Cohan et al.\\ (2019) \cite{Cohan2019PretrainedLM} \end{tabular}                & SciBERT                                                          & \begin{tabular}[c]{@{}l@{}}SciBERT-\\$[SEP]$\end{tabular}                                                             & \begin{tabular}[c]{@{}l@{}}SciBERT-\\$[SEP]$\end{tabular}                                                 & softmax                                                    \\ \hline
\begin{tabular}[c]{@{}l@{}}Gon{\c{c}}alves et al.\\ (2020) \cite{Goncalves20} \end{tabular}     & GloVe                                                            & \begin{tabular}[c]{@{}l@{}}CNN / \\ max pooling\end{tabular}           & Bi-GRU                                                        & softmax                                                    \\ \hline
\begin{tabular}[c]{@{}l@{}}Yamada et al.\\ (2020) \cite{yamada-etal-2020-sequential}\end{tabular}              & \begin{tabular}[c]{@{}l@{}}BERT from \\ PubMed\end{tabular}      & \begin{tabular}[c]{@{}l@{}}BERT-\\$[CLS]$\end{tabular}                                                                & Bi-LSTM                                                       & \begin{tabular}[c]{@{}l@{}}Semi-\\Markov CRF\end{tabular} \\ \hline
\begin{tabular}[c]{@{}l@{}}Shang et al.\\ (2021) \cite{Shang21}\end{tabular}              & \begin{tabular}[c]{@{}l@{}}SciBERT\end{tabular}      & \begin{tabular}[c]{@{}l@{}}SciBERT-\\$[CLS]$\end{tabular}                                                                & \begin{tabular}[c]{@{}l@{}}Bi-LSTM/\\attention\end{tabular}                                                       & \begin{tabular}[c]{@{}l@{}}CRF\end{tabular} \\

\end{tabular}
\end{table}

\paragraph{Approaches for Full Papers:}
For full papers, logistic regression, support vector machines and CRFs with hand-crafted features have been proposed \citep{Asadi2019AutomaticZI,Badie2018ZoneIB,Fisas2015OnTD,liakata2012automatic,teufel1999argumentative,teufel2009towards}. 
They represent a sentence with various syntactic and linguistic features such as n-grams, part-of-speech tags, or citation markers, which were engineered for the respective datasets. 
\citet{Asadi2019AutomaticZI} also exploit semantic features obtained from knowledge bases such as Wordnet~\citep{Fellbaum2000WordNetA}. 
To incorporate contextual information, each sentence representation also contains the label of the previous sentence (``history feature'') and the sentence position in the document (``location feature'').
To better consider the interdependence between labels, some approaches apply CRFs, while \citet{Asadi2019AutomaticZI} suggest fusion techniques within a dynamic window of sentences.
However, some approaches \cite{Asadi2019AutomaticZI,Badie2018ZoneIB,Fisas2015OnTD} exploit the \textit{ground-truth label} instead of the predicted label of the preceding sentence (``history feature'') during prediction (as confirmed by the authors), which has a significant impact on the performance. 

Related tasks also classify sentences in full papers with deep learning methods, 
e.g.\ for citation intent classification~\citep{Cohan2019StructuralSF,kunnath-etal-2020-overview},
or algorithmic metadata extraction~\citep{Safder2020} but without exploiting context from surrounding sentences.
Comparable to us, \citet{lauscher2018a} utilise a hierarchical deep learning architecture for argumentation mining in full papers but evaluate it only on one corpus.

\textit{To the best of our knowledge, a unified approach for sequential sentence classification for abstracts as well as full papers has not been proposed and evaluated yet}.

\subsection{Transfer Learning}
Transfer learning enables a target task to exploit knowledge from another source task to achieve a better prediction accuracy. 
The tasks can have training data from different domains and vary in their objectives.
According to Ruder's taxonomy for transfer learning ~\citep{Ruder2019Neural}, we investigate inductive transfer learning in this study since the target training datasets are labelled.
Inductive transfer learning can be further subdivided into multi-task learning, where tasks are learned simultaneously, and sequential transfer learning (also referred to as parameter initialisation), where tasks are learned sequentially.
Since there are many applications for transfer learning, we focus on the most relevant cases to our work here.
For a more comprehensive overview, we refer to \cite{Pan10,Ruder2019Neural,Weiss16TL}.

\emph{Fine-tuning a pre-trained language model} is a popular approach for sequential transfer learning in NLP~\cite{Brown20,Devlin2018BERTPO,He21,Ruder18}.
Here, the source task involves learning a language model (or a variant of it) using a large unlabelled text corpus. Then, the model parameters are fine-tuned with labelled data of the target task. 
\citet{Pruksachatkun20} improve these language models by \emph{intermediate task transfer learning} where a language model is fine-tuned on a data-rich intermediate task before fine-tuning on the final target task.
\citet{park-caragea-2020-scientific} provide an empirical study on intermediate transfer learning from the non-academic domain to scientific keyphrase identification. They show that SciBERT in combination with related tasks such as sequence tagging improves performance, while BERT or unrelated tasks degrade the performance.

For \emph{sequence tagging}
Yang et al.~\cite{Yang2017TransferLF} investigate multi-task learning in the general non-academic domain with a small and a big dataset.
\citet{SchulzEDKG18} evaluate multi-task learning for argumentation mining with multiple datasets in the general domain.
\citet{Lee2017TransferLF} successfully transfer pre-trained parameters from a big dataset to a small dataset in the biological domain. 
For \emph{coreference resolution}, \citet{Brack2021Coref} apply sequential transfer learning and utilise a large dataset from the general domain to improve models for a small dataset in the scientific domain.

For \emph{sentence classification},
\citet{Mou2016HowTA} compare (1) transferring parameters from a source dataset to a target dataset against (2) training one model with two datasets in the non-academic domain. They demonstrate that semantically related tasks improve while unrelated tasks degrade the performance of the target tasks. 
\citet{Su20} study multi-task learning for sentiment classification in product reviews from multiple domains.
\citet{LauscherGP018} evaluate multi-task learning on scientific texts, however, only on one dataset with different annotation layers.
\citet{Banerjee20} apply sequential transfer learning from the medical to the computer science domain for discourse classification, however, only for two domains and on abstracts, whereas \citet{Spangher21}
explore this task on news articles with multi-task learning using multiple datasets.
\citet{Gupta21} utilise a multi-task learning with two scaffold tasks to detect contribution sentences in full papers, however, only in one domain and with limited sentence context.

Several approaches also exist to \emph{train multiple tasks jointly}:
\citet{Luan2018MultiTaskIO} train a model on three tasks (coreference resolution, entity and relation extraction) using one dataset of research papers. 
\citet{Wei2019JointEO} utilise a multi-task model for entity recognition and relation extraction on one dataset in the non-academic domain.
\citet{Changpinyo18} analyse multi-task training with multiple datasets for sequence tagging. 
\textit{In contrast, we investigate sequential sentence classification across multiple science domains.}

\section{Sequential Sentence Classification}

On the one hand, the discussion of related work shows that several approaches and datasets from various scientific domains have been introduced for sequential sentence classification. 
On the other hand, although transfer learning has been applied to various NLP tasks, it is known that the success depends largely on the relatedness of the tasks~\citep{Mou2016HowTA,Pan10,Ruder2019Neural}. 
However, the field lacks an empirical study on transfer learning between different scientific domains for sequential sentence classification that cover either only abstracts or entire papers.
Furthermore, previous approaches investigated transfer learning for one or two datasets only.
To the best of our knowledge, a unified approach for different types of texts that differ noticeably by their structure and semantic context of sentences, as it is the case for abstracts and full papers, has not been proposed yet.

\begin{figure*}[t]
    \center{\includegraphics[width=0.8\linewidth]
        {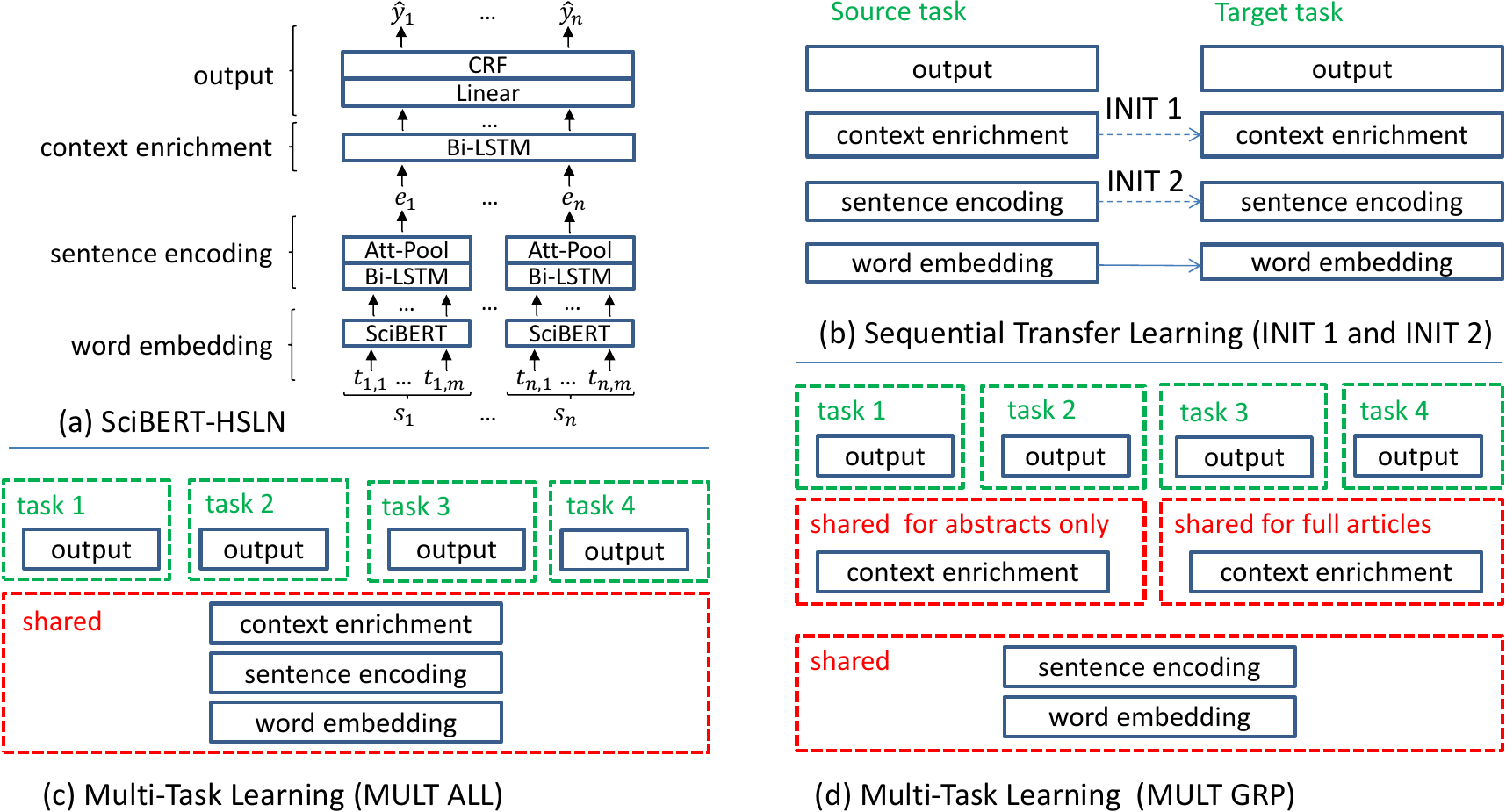}}
    \caption{Proposed approaches for sequential sentence classification: (a) unified deep learning architecture \emph{SciBERT-HSLN} for datasets of 
    abstracts and full papers; (b) sequential transfer learning approaches, i.e.\ INIT 1 transfers all possible layers, INIT 2 only the sentence encoding layer; (c) and (d) are the multi-task learning approaches, i.e.\ in MULT ALL all possible layers are shared between the tasks, in MULT GRP the context enrichment is shared between tasks with the same text type.}
    \label{fig:proposed_approach}
\end{figure*}

In this section, we suggest a unified cross-domain multi-task learning approach for sequential sentence classification.
Our tailored transfer learning approaches, depicted in Figure~\ref{fig:proposed_approach}, exploit multiple datasets comprising different text types in form of abstracts and full papers. The unified approach without transfer learning is described in Section~\ref{sec:unified_approach} while Section~\ref{sec:transfer_methods} introduces our tailored transfer learning approaches.
Finally, in Section~\ref{sec:sem_relatedness}, we present an approach to semi-automatically identify the semantic relatedness of sentence classes between different annotation schemes.

\subsection{Unified Deep Learning Approach}
\label{sec:unified_approach}
Given a paper with the sentences ($s_1, ..., s_n$) and the set of dataset specific classes $L$ (e.g.\ \emph{Background}, \emph{Methods}), the task of \emph{sequential sentence classification} is to predict the corresponding label sequence ($y_1, ..., y_n$) with $y_i \in L$.
For this task, we propose a unified deep learning approach as depicted in Figure~\ref{fig:proposed_approach}(a), which is applicable to both abstracts \emph{and} full papers. 
The core idea is to enrich sentence representations with context from surrounding sentences..

Our approach (denoted as \emph{SciBERT-HSLN}) is based on the \emph{Hierarchical Sequential Labeling Network (HSLN)} \cite{Jin2018HierarchicalNN}. In contrast to \citet{Jin2018HierarchicalNN}, we utilise SciBERT~\citep{Beltagy2019SciBERTPC} as word embeddings and evaluate the approach on abstracts \emph{as well as} full papers. 
We have chosen HSLN as the basis since it is better suited for full papers: It has no limitations on text length (in contrast to the approach of \citet{Cohan2019PretrainedLM}), and is computationally less expensive than the more recent approaches~\cite{Shang21,yamada-etal-2020-sequential}. 
Furthermore, their implementation is publicly available.
The goal of this paper is not to beat state-of-the-art results but rather to provide an empirical study on transfer learning and offer a uniform solution.
Our \emph{SciBERT-HSLN} architecture has the following layers:

(a)~\emph{Word embedding:} input is a sequence of tokens $(t_{i,1}, ..., t_{i,m})$ of sentence $s_i$, and output a sequence of (SciBERT) word embeddings $(w_{i,1}, ..., w_{i,m})$. 

(b)~\emph{Sentence encoding:} input  $(w_{i,1}, ..., w_{i,m})$ is transformed via a Bi-LSTM~\cite{Hochreiter1997LongSM} into the representations $(h_{i,1}, ..., h_{i,m})$ ($h_{i,t} \in \mathbb{R}^{d^h}$) which are enriched with contextual information within the sentence.
Then, attention pooling~\citep{Jin2018HierarchicalNN,Yang2016HierarchicalAN} with $r$ heads produces a sentence vector $e_i \in \mathbb{R}^{rd^u}$.
An attention head produces a weighted average over the token representations of a sentence.
Multiple heads enable the model to capture several semantics of a sentence. 

(c)~\emph{Context enrichment:} sentence vectors $(e_1,...,e_n)$ are transformed via a Bi-LSTM into $(c_1,...,c_n)$ with $c_i \in \mathbb{R}^{d^h}$. Thus, each sentence vector $c_i$ is enriched with contextual information from surrounding sentences. 

(d)~\emph{Output layer:} transforms $(c_1,...,c_n)$ via a linear transformation to the logits $(l_1, ..., l_n)$ with $l_i \in \mathbb{R}^{|L|}$.
Each component in $l_i$ contains a score for the corresponding label. 
A CRF~\cite{Lafferty2001ConditionalRF} predicts the labels $(\hat y_1, ..., \hat y_n)$ with $\hat y_i \in L$ with the highest conditional joint probability $P(\hat y_1, ..., \hat y_n|l_1, ..., l_n)$.
In this way, it makes use of patterns that appear in scientific papers (e.g.\ \textit{Methods} are usually followed by \textit{Results}).
During training the CRF maximises $P(y_1, ..., y_n|l_1, ..., l_n)$ of the ground-truth labels for all training samples. 
The Viterbi algorithm~\cite{Forney73viterbi} is used for efficient prediction and training.

For regularisation, we use dropout after each layer.
We do not fine-tune Sci\-BERT embeddings, since it requires training of 110 Mio. additional parameters.

\subsection{Transfer Learning Methods}
\label{sec:transfer_methods}
For sequential sentence classification, we tailor and evaluate the following transfer learning methods.

\paragraph{Sequential Transfer Learning (INIT):}
The approach first trains the model for the source task and uses its tuned parameters to initialise the parameters for the target task. 
Then, the parameters are fine-tuned with the labelled data of the target task. As depicted in Figure~\ref{fig:proposed_approach}(b), we propose two types of layer transfers. 
\emph{INIT~1}: transfer parameters of \emph{context enrichment} and \emph{sentence encoding}; 
\emph{INIT~2}: transfer parameters of \emph{sentence encoding}. Other layers, except \emph{word embedding}, of the target task are initialised with random values. 

\paragraph{Multi-Task Learning (MULT):}
Multi-task learning (MULT) aims for a better generalisation by simultaneously training samples in all tasks and sharing parameters of certain layers between the tasks. 
As depicted in Figure~\ref{fig:proposed_approach}(c,d), we propose two multi-task learning architectures.
The \emph{MULT ALL} model shares all layers between the tasks except the \emph{output layers} so that the model learns a common feature extractor for all tasks.
However, full papers are much longer and have a different rhetorical structure compared to abstracts. 
Therefore, it is not beneficial to share the context enrichment layer between both dataset types. Thus, in the \emph{MULT GRP} model, the \emph{context enrichment layers} are only shared between datasets with the same text type.
Formally, the objective is to minimise the following loss functions: 
\begin{eqnarray}
\textstyle L_{\mbox{\tiny MULT ALL}} = \sum_{t \in T^A \cup T^F} L_t(\Theta^S, \Theta^C, \Theta^O_{t})  \\
\begin{aligned}
\textstyle L_{\mbox{\tiny MULT GRP}} = \sum_{t \in T^A} L_t(\Theta^S, \Theta^{C^A}, \Theta^O_{t})  
+ \sum_{t \in T^F} L_t(\Theta^S, \Theta^{C^F}, \Theta^O_{t}) 
\end{aligned}
\end{eqnarray}
where $T^A$ and $T^F$ are the tasks for datasets containing abstracts and full papers;
$L_t$ is the loss function for task $t$;
the parameters 
$\Theta^S$ are for sentence encoding, 
$\Theta^C$, $\Theta^{C^A}$, $\Theta^{C^F}$ for context enrichment,
and $\Theta^{O}_{t}$ for the output layer of task $t$.

Furthermore, we propose the variants \emph{MULT ALL SHO} and \emph{MULT GRP SHO} that are applicable if all tasks share the same (domain-independent) set of classes. 
\emph{MULT  ALL  SHO} shares all layers among all tasks. \emph{MULT  GRP SHO} shares the context enrichment and output layer only between tasks with the same text type. Formally, the objective functions are defined as:
\begin{eqnarray}
\textstyle L_{\mbox{\tiny MULT ALL SHO}} = \sum_{t \in T^A \cup T^F} L_t(\Theta^S, \Theta^C, \Theta^O)  \\
\begin{aligned}
\textstyle L_{\mbox{\tiny MULT GRP SHO}} = \sum_{t \in T^A} L_t(\Theta^S, \Theta^{C^A}, \Theta^{O^A})  
\\ 
\textstyle + \sum_{t \in T^F} L_t(\Theta^S, \Theta^{C^F}, \Theta^{O^F}) 
\end{aligned}
\end{eqnarray}

\subsection{Semantic Relatedness of Classes}
\label{sec:sem_relatedness}

Datasets for sentence classification have different domain-specific annotation schemes, that is different sets of pre-defined classes.
Intuitively, some classes have a similar meaning across domains, e.g.\ the classes \textit{Model} and \textit{Experiment} in the ART corpus are semantically related to \textit{Methods} in PubMed-20k (PMD) (see Table~\ref{table:datasets}).
An analysis of semantic relatedness can help consolidate different annotation schemes.
We propose machine learning models to support the identification of semantically related classes according to the following idea: If a model trained for PMD recognises sentences labelled with \textit{ART:Model} as \textit{PMD:Method}, and vice versa, then the classes \textit{ART:Model} and \textit{PMD:Method} can be assumed to be semantically related.

Let $T$ be the set of all tasks, 
$L$ the set of all classes in all tasks,
$m_t(s)$ the label of sentence $s$ predicted by the model for task $t$, and
$S^l$ the set of sentences with the ground truth label $l$.
For each class $l \in L$ the corresponding semantic vector $v_l \in \mathbb{R}^{|L|}$ is defined as:
\begin{eqnarray}
v_{l,l'} = \frac{\sum_{t \in T, s \in S^l} {\mathbbm{1}(m_t(s) = l')}}{|S^l|}
\end{eqnarray}
where $v_{l, l'} \in \mathbb{R}$ is the component of the vector $v_l$ for class $l' \in L$ and 
$\mathbbm{1}(p)$ 
is the indicator function that returns $1$ if $p$ is true and $0$ otherwise.
Intuitively, the semantic vectors concatenated vertically to a matrix represent a ``confusion matrix'' (see Figure~\ref{fig:pred_heatmap} as an example). 
Now, we define the semantic relatedness of two classes $k, l \in L$ using cosine similarity:
\begin{eqnarray}
\text{semantic\_relatedness}(k, l) = \cos(v_{k}, v_{l}) = \frac{v_{k}^\intercal \cdot v_{l}}{||v_{k}|| \cdot ||v_{l}||} \label{eq:semantic_relatedness}
\end{eqnarray}

\section{Experimental Setup}

This section describes the experimental evaluation of the proposed approaches, i.e.\ used datasets, implementation details, and evaluation methods.

\begin{table}[tb]
\centering
\footnotesize
\caption{Characteristics of the benchmark datasets. The row "SOTA" depicts the best results for approaches that do not exploit the ground-truth label of the preceding sentence during prediction: for PMD \protect\citep{yamada-etal-2020-sequential}, for NIC \protect\citep{Shang21}, 
for DRI \protect\citep{Badie2018ZoneIB} (cf. Table 7), and for ART \protect\citep{liakata2012automatic}.}
\begin{tabular}{l|r|r|r|r}
             & \textbf{PMD}   & \textbf{NIC}       & \textbf{DRI} & \textbf{ART}       \\
\hline
Domains             & Biomedicine  & Biomedicine  & Computer &  Chemistry, \\
                   &              &              & Graphics &  Comp.          \\
                   &              &              &          &  Linguistic          \\
\hline
Text Type &  Abstract   & Abstract & Full article & Full article           \\
\hline
\# Articles &  20.000   & 1.000 & 40 & 225           \\
\# Sentences &  235.892    & 9.771  & 8.777 & 34.680           \\
$\varnothing$ \# Sent.  & 12  & 10   & 219  &  154          \\
\# Classes &  5    & 6  & 5 & 11           \\
\hline
Classes & Background     & Background & Background & Background                        \\
        & Objective & Intervention & Challenge & Motivation                        \\
        & Methods & Study & Approach                                   & Hypothesis                        \\
        & Results & Population & Outcome                                    & Goal       \\
        & Conclusion & Outcome & FutureWork                                 & Object                       \\
        &                         & Other                                      &                                            & Experiment                        \\
        &                                      &                                            &                                            & Model                             \\
        &                                      &                                            &                                            & Method                            \\
        &                                      &                                            &                                            & Observation                            \\
        &                                      &                                            &                                            & Result                            \\
        &                                      &                                            &                                            & Conclusion                        \\ 
\hline

SOTA   & \cite{yamada-etal-2020-sequential} \underline{93.1}       & \cite{Shang21} \underline{86.8}  & \cite{Badie2018ZoneIB} 72.5  & \cite{liakata2012automatic} 51.6\\ 
metric   & weighted F1 & weighted F1 & weighted F1 & accuracy \\  

\end{tabular}
\label{table:datasets}
\end{table}

\subsection{Investigated Datasets}
Table~\ref{table:datasets} summarises the characteristics of the investigated datasets, namely PubMed-20k (PMD)~\citep{Dernoncourt2017PubMed2R}, NICTA-PIBOSO (NIC) \citep{Kim2011AutomaticCO}, ART \citep{Liakata2010CorporaFT}, and Dr. Inventor (DRI)~\citep{Fisas2015OnTD}. 
The four datasets are publicly available and provide a good mix to investigate the transferability: They represent four different scientific domains; PMD and NIC cover abstracts and are from the same domain but have different annotation schemes; DRI and ART cover full papers but are from different domains and have different annotation schemes; NIC and DRI are rather small datasets, while PMD and ART are about 20 and 3 times larger, respectively; ART has a much finer annotation scheme compared to other datasets.
As denoted in Table~\ref{table:datasets}, the state-of-the-art results for ART are the lowest ones since ART has more fine-grained classes than the other datasets. 
In contrast, best results are obtained for PMD: It is a large dataset sampled from PubMed, where authors are encouraged to structure their abstracts. 
Therefore, abstracts in PMD are more uniformly structured than in other datasets, leading to better classification results.

\subsection{Implementation}
\label{sec:implementation}
Our approaches are implemented in PyTorch~\citep{Ketkar2017IntroductionTP}.
The Adaptive Moment Estimation (ADAM) optimiser~\citep{Kingma2014AdamAM} with $0.01$ weight decay and an exponential learning rate decay of $0.9$ after each epoch is used for training. 
To speed up training, sentences longer than 128 tokens are truncated since the computational cost for the attention layers in BERT is quadratic in sentence length~\citep{VaswaniSPUJGKP17}.
To reproduce the results of the original HSLN architecture, we tuned SciBERT-HSLN for PMD and NIC with hyperparameters as proposed in other studies ~\citep{Devlin2018BERTPO,Jin2018HierarchicalNN}.
The following parameters performed best on the validation sets of PMD and NIC: learning rate  3e-5, dropout rate $0.5$, Bi-LSTM hidden size $d^h=2 \cdot 758$, $r=15$ attention heads of size $d^u=200$. 
We used these hyperparameters in all of our experiments.

For each dataset, we grouped papers to mini-batches without splitting them, if the mini-batch does not exceed 32 sentences. Thus, for full papers a mini-batch may consist of sentences from only one paper.
During multi-task training we switched between the mini-batches of the tasks by 
proportional sampling~\citep{Sanh2018AHM}.
After a mini-batch, only task-related parameters are updated, i.e.\ the associated output layer and all the layers below.

\subsection{Evaluation}
To be consistent with previous results and due to non-determinism in deep neural networks~\citep{Reimers2017ReportingSD}, we repeated the experiments and averaged the results.
According to \citet{Cohan2019PretrainedLM} we performed three random restarts for PMD and NIC and used the same train/valida\-tion/test sets. 
For DRI and ART, we performed 10-fold and 9-fold cross-validation, respectively, as in the original papers \citep{Fisas2015OnTD,liakata2012automatic}. 
Within each fold the data is split into train/validation/test sets with the proportions $\frac{k-2}{k}$/$\frac{1}{k}$/$\frac{1}{k}$ where $k$ is the number of folds. 
For multi-task learning, the experiment was repeated with the maximum number of folds of the datasets used, but at least three times.
All models were trained for 20 epochs. The test set performance within a fold and restart, respectively, was calculated for the epoch with the best validation performance. 

We compare our results only with approaches which do not exploit \emph{ground-truth labels} of the preceding sentence as a feature \emph{during prediction} (see Section~\ref{sec:ml_approaches}). 
This has a significant impact on the performance: Using the ground truth label of the previous sentences as a sole input feature to a SVM classifier already yields an accuracy of 77.7 for DRI and 55.5 for ART (compare also results for the ``history'' feature in~\cite{Badie2018ZoneIB}, cf.\ Table 5).
Best reported results using ground truth labels as input features have an accuracy of 84.15 for DRI and 65.75 for ART~\citep{Asadi2019AutomaticZI}. 
In contrast, we pursue a realistic setting by exploiting the \emph{predicted} (not ground truth) label of neighbouring sentences during prediction.

Moreover, we provide additional results for three strong deep learning baselines:
(1) fine-tuning SciBERT using the [CLS] token of individual sentences as in \cite{Devlin2018BERTPO} (referred to as SciBERT-[CLS]),
(2) original HSLN implementation of \citet{Jin2018HierarchicalNN}, and
(3) the SciBERT-based approach of \citet{Cohan2019PretrainedLM}.
We cannot provide baseline results for DRI and ART of the approaches~\cite{Shang21,yamada-etal-2020-sequential} since their implementations are not publicly available.

\begin{table}[t!]
\centering
\footnotesize
\caption{Experimental results for the proposed approaches: our SciBERT-HSLN model without transfer learning, parameter initialisation (INIT), and multi-task learning (MULT ALL and MULT GRP). Previous state of the art (see Table~\ref{table:datasets}), SciBERT-[CLS], and the approaches of \citet{Jin2018HierarchicalNN} and \citet{Cohan2019PretrainedLM} are the baseline results. For PMD (P), NIC (N), and DRI (D) we report weighted F1 score and for ART (A) accuracy. The average of all scores is denoted by $\varnothing$. \textit{Italics} depicts whether the result is better than the baseline, \textbf{bold} whether the transfer method improves SciBERT-HSLN, \underline{underline} the best overall result.}
\begin{tabular}{l|r|r|r|r|r}
                                           & \textbf{   PMD    } & \textbf{    NIC    } & \textbf{    DRI    }   & \textbf{    ART    }   & $\varnothing$ \\ 
                                           \hline
\textbf{Prev. SOTA}                                       & \cite{yamada-etal-2020-sequential} \underline{93.1}       & \cite{Shang21} \underline{86.8}  & \cite{Badie2018ZoneIB} 72.5  & \cite{liakata2012automatic} 51.6 & 76.0 \\ 
SciBERT-[CLS]~\citep{Beltagy2019SciBERTPC}                                       & 89.6       & 78.4  & 69.5  & 51.5 & 72.3 \\ 
\citet{Jin2018HierarchicalNN}                                & 92.6       & 84.7  & 75.3  & 49.3  & 75.5  \\
\citet{Cohan2019PretrainedLM}           & 92.9       & 84.8  & 74.3  & 54.3 & 76.6 \\ 
\hline
\hline
\textbf{SciBERT-HSLN}                                & 92.9      & \textit{84.9} & \textit{78.0} & \textit{58.0} & \textit{78.5} \\ 
\hline
\hline
INIT 1 PMD to $T$                           & -          & 84.8 & \textbf{\textit{81.2}} & \textit{57.7} \\
INIT 2 PMD to $T$                           & -          & 84.8 & \textbf{\textit{80.1}} & \textit{58.0} \\
\hline
INIT 1 NIC to $T$                           &           92.9  &- & \textbf{\textit{81.9}}  &  \textit{57.6} \\
INIT 2 NIC to $T$                           &           92.9  &- & \textbf{\textit{79.6}}  &  \textit{57.2} \\
\hline
INIT 1 DRI to $T$                           &           92.9 & 83.5 & -  &  \textit{57.8} \\
INIT 2 DRI to $T$                           &           92.9 & 83.8 & -  &  \textit{57.6} \\
\hline
INIT 1 ART to $T$                           &           \textbf{93.0} & 84.7 & \textbf{\textit{82.2}}  &  - \\
INIT 2 ART to $T$                           &           92.9 & 84.7 & \textbf{\textit{81.0}}  &  - \\
\hline
\hline
\textbf{MULT ALL}                                   & \textbf{93.0}      & \textbf{86.0} & \textbf{\textit{81.8}} & \textit{57.7} & \textbf{\textit{79.6}}\\
\hline

PMD, NIC                         & \textbf{93.0}      & \textbf{86.1} & -     & -     \\
PMD, DRI                           & 92.9      & -     & \textbf{\textit{80.6}}  & -     \\
PMD, ART                           & \textbf{93.0}      & -     & -     & \textit{58.0} \\

NIC, DRI                          & -          & 84.2  & \textbf{\textit{80.7}} & -     \\
NIC, ART                          & -          & 84.4 & -     & \textit{57.9} \\
DRI, ART                            & -          & -     & \textbf{\textit{82.0}}    & \textit{57.6} \\

\hline

PMD, NIC, DRI                    & \textbf{93.0}      & \textbf{86.2} & \textbf{\textit{81.0}} & -     \\
PMD, NIC, ART                     & \textbf{93.0}          & \textbf{86.3}  & - & \textit{58.0} \\
PMD, DRI, ART                     & \textbf{93.0}          &   -& \textbf{\textit{82.7}} & \textit{57.8} \\
NIC, DRI, ART                     & -          & 84.7 & \textbf{\textit{82.0}} & \textit{57.7} \\

\hline
\hline
\textbf{MULT GRP}  & \textbf{93.0}  & \textbf{86.1} & \textbf{\textit{83.4}} & \underline{\textbf{\textit{58.8}}} & \underline{\textbf{\textit{80.3}}} \\
\hline
 P,N,D,A  & 92.9  & \textbf{85.4} &  \underline{\textbf{\textit{84.4}}} & \textit{58.0} & \textbf{\textit{80.2}}\\ 
\hline
(P,D),(N,A)  & \textbf{93.0}  & \textbf{86.0} & \textbf{\textit{81.1}} & \textbf{\textit{58.5}} & \textbf{\textit{79.7}} \\ 
(P,A),(N,D)  & 92.9  & \textbf{85.8} & \textbf{\textit{83.6}} & \textit{58.0} & \textbf{\textit{80.1}} \\ 
\hline

 (P,N,D),(A)  & 92.9  & \textbf{86.0} & \textbf{\textit{80.6}} & \textbf{\textit{58.2}} & \textbf{\textit{79.4}} \\ 
 (P,N,A),(D)  & \textbf{93.0}  & \textbf{86.0} & \textbf{\textit{84.1}} & \textbf{\textit{58.1}} & \textbf{\textit{80.3}} \\ 
(P,D,A),(N)  & 92.9  & \textbf{85.5} & \textbf{\textit{82.2}} & \textit{58.0} & \textbf{\textit{79.6}} \\ 
 (N,D,A),(P)  & 92.9  & \textbf{85.9} & \textbf{\textit{83.3}} & \textbf{\textit{58.5}} & \textbf{\textit{80.1}} \\

\end{tabular}
\label{table:results}
\end{table}

\section{Results and Discussion}
In this section, we present and discuss the experimental results for our proposed cross-domain multi-task learning approach for sequential sentence classification. 
The results for different variations of our approach, the respective baselines, and for several state-of-the-art methods are depicted in Table~\ref{table:results}.
The results are discussed in the following three subsections with regard to the unified approach without transfer learning (Section \ref{sec:notransfer}), with sequential transfer learning (Section \ref{sec:transferinit}), and multi-task learning (Section \ref{sec:transfermulti}).
Section~\ref{sec:analysis_semantic_relatedness} analyses the semantic relatedness of classes for the four annotation schemes.

\subsection{Unified Approach without Transfer Learning (SciBERT-HSLN)}
\label{sec:notransfer}

For the full paper datasets DRI and ART, our SciBERT-HSLN model significantly outperforms the previously reported best results, and the deep learning baselines SciBERT-[CLS], \citet{Jin2018HierarchicalNN}, and \citet{Cohan2019PretrainedLM}.
The previous state of the art approaches for DRI and ART~\citep{Badie2018ZoneIB,liakata2012automatic} require feature engineering and a sentence is enriched only with the context of the previous sentence. In SciBERT-[CLS], each sentence is classified in isolation.
The original HSLN architecture \citep{Jin2018HierarchicalNN} uses shallow word embeddings pre-trained on biomedical texts. 
Thus, the incorporation of SciBERT's contextual word embeddings into HSLN helps improve performance for the DRI and ART datasets.
The approach of \citet{Cohan2019PretrainedLM} can process only about 10 sentences at once since SciBERT supports sequences of up to 512 tokens only. Thus, long text has to be split into multiple chunks. Our deep learning approach can process \emph{all} sentences of a paper at once so that all sentences are enriched with context from surrounding sentences.

For the PMD dataset, our SciBERT-HSLN results are equivalent~\cite{yamada-etal-2020-sequential} to the current state of the art, while for NIC, they are below~\cite{Shang21}.
Thus, our proposed approach is competitive with the current approaches for sequential sentence classification in abstracts.
\emph{Our unified deep learning approach is applicable to datasets consisting of different text types, i.e.\ abstracts and full papers, without any feature engineering (RQ7).}

\subsection{Sequential Transfer Learning (INIT)}
\label{sec:transferinit}
Using the INIT approach, we can only improve the baseline results for the DRI dataset in all settings. 
The approach INIT~1 performs better than INIT~2 in most cases which indicates that transferring all parameters is more effective.
\emph{However, the results suggest that sequential transfer learning is not a very effective transfer method for sequential sentence classification (RQ2).}

\begin{table}[t]
\centering
\footnotesize
\caption{Experimental results for $\mu$PMD, NIC, DRI and $\mu$ART with our SciBERT-HSLN model and our proposed multi-task learning approaches.}
\begin{tabular}{l|r|r|r|r|r}
         & \textbf{$\mu$PMD} & \textbf{NIC} & \textbf{DRI}   & \textbf{$\mu$ART} & $\varnothing$  \\
\hline         
SciBERT-HSLN & 90.9  & 84.9 & 78.0 & 52.2 & 76.5\\ 
MULT ALL & \textbf{91.1}  & \textbf{85.7} & \textbf{81.0} & \textbf{53.8} & \textbf{77.9} \\
MULT GRP   & \textbf{91.1}  & \textbf{85.9} & \textbf{82.2} & \textbf{55.1} & \textbf{78.6}  
\end{tabular}
\label{table:small_results}
\end{table}

\begin{figure}[t]
    \center{\includegraphics[width=0.85\linewidth]
        {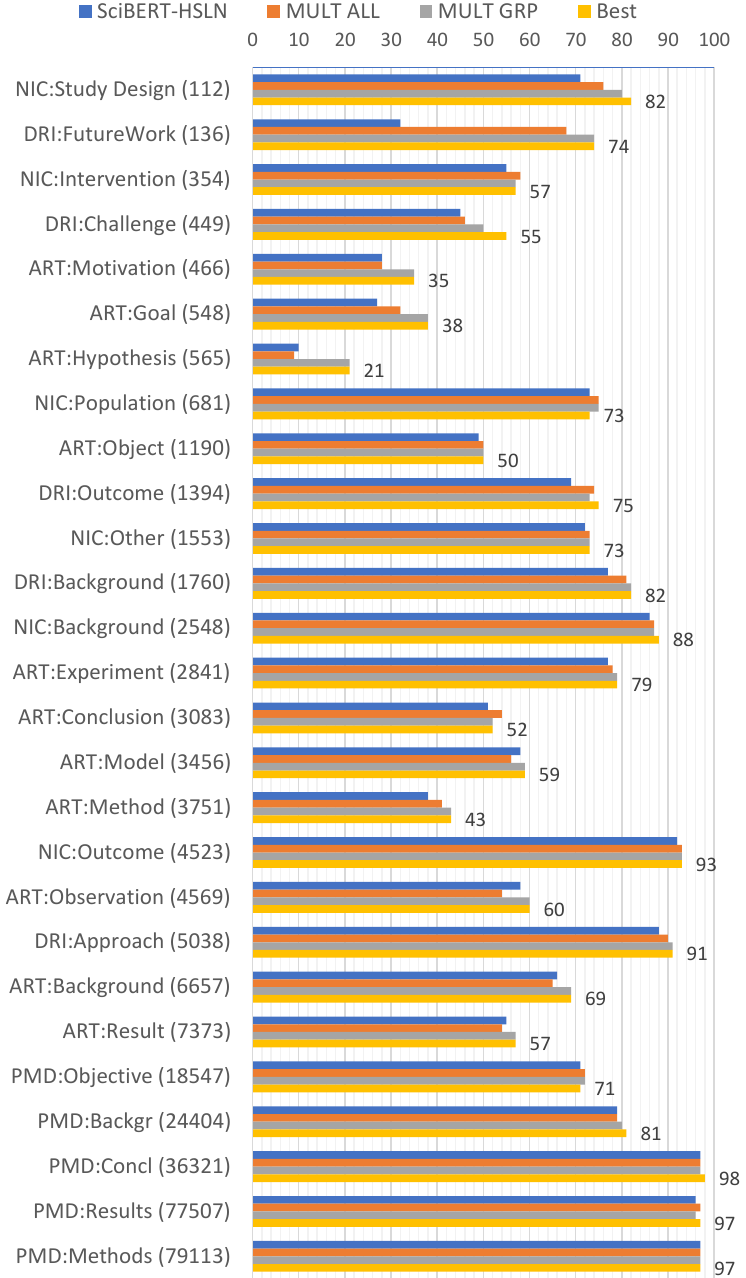}}
    \caption{F1 scores per class for the datasets PMD, NIC, DRI, and ART for SciBERT-HSLN, MULT ALL, MULT GRP, and the best combination for the respective dataset. Numbers at the bars depict the F1 scores of the best classifiers and in brackets the number of examples for the given class. The classes are ordered by the number of examples.}
    \label{fig:f1_per_label}
\end{figure}

\subsection{Multi-Task Learning (MULT)}
\label{sec:transfermulti}
Next, we discuss the results of our multi-task learning approach, and the effects of multi-task learning on smaller datasets and individual sentence classes.

\paragraph{MULT ALL model:}
All tasks were trained jointly sharing all possible layers. 
Except for the ART task, all results are improved using the SciBERT-HSLN model.
For the PMD task, the improvement is marginal since the baseline results (F1 score) were already on a high level.
Pairwise MULT ALL combinations show that the models for PMD and NIC, respectively, benefit from the (respective) other dataset, and the DRI model especially from the ART dataset. 
The PMD and NIC datasets are from the same domain, and both contain abstracts, so the results are as expected. Furthermore, DRI and ART datasets both contain full papers, and DRI has more coarse-grained classes. However, ART is a related large dataset with fine-grained classes and presumably therefore the model for ART does not benefit from other datasets.
In triple-wise MULT ALL combinations the models for PMD and DRI, respectively, benefit from all datasets, and the model for NIC only if the PMD dataset is present. 
\emph{The results suggest that sharing all possible layers between multiple tasks is effective except for bigger datasets with more fine-grained classes (RQ3, RQ4).}

\paragraph{MULT GRP model:}
In this setting, the models for all tasks were trained jointly, but only models for the same text type share the \emph{context enrichment layer}, i.e.\ (PMD, NIC) and (DRI, ART). 
Here, all models benefit from the other datasets. 
In our ablation study, we also provide results for sharing only the \emph{sentence encoding layer}, referred to as MULT GRP P,N,D,A, and all pairwise and triple-wise combinations sharing the \emph{context enrichment layer}.
Other combinations also yield good results. However, MULT GRP is effective for \emph{all} tasks. 
\emph{Our results indicate that sharing the sentence encoding layer between multiple models is beneficial. Furthermore, sharing the context enrichment layer only between models for the same text type is an even more effective strategy (RQ3, RQ4).}

\begin{figure*}[t]
    \center{\includegraphics[width=0.8\linewidth]
        {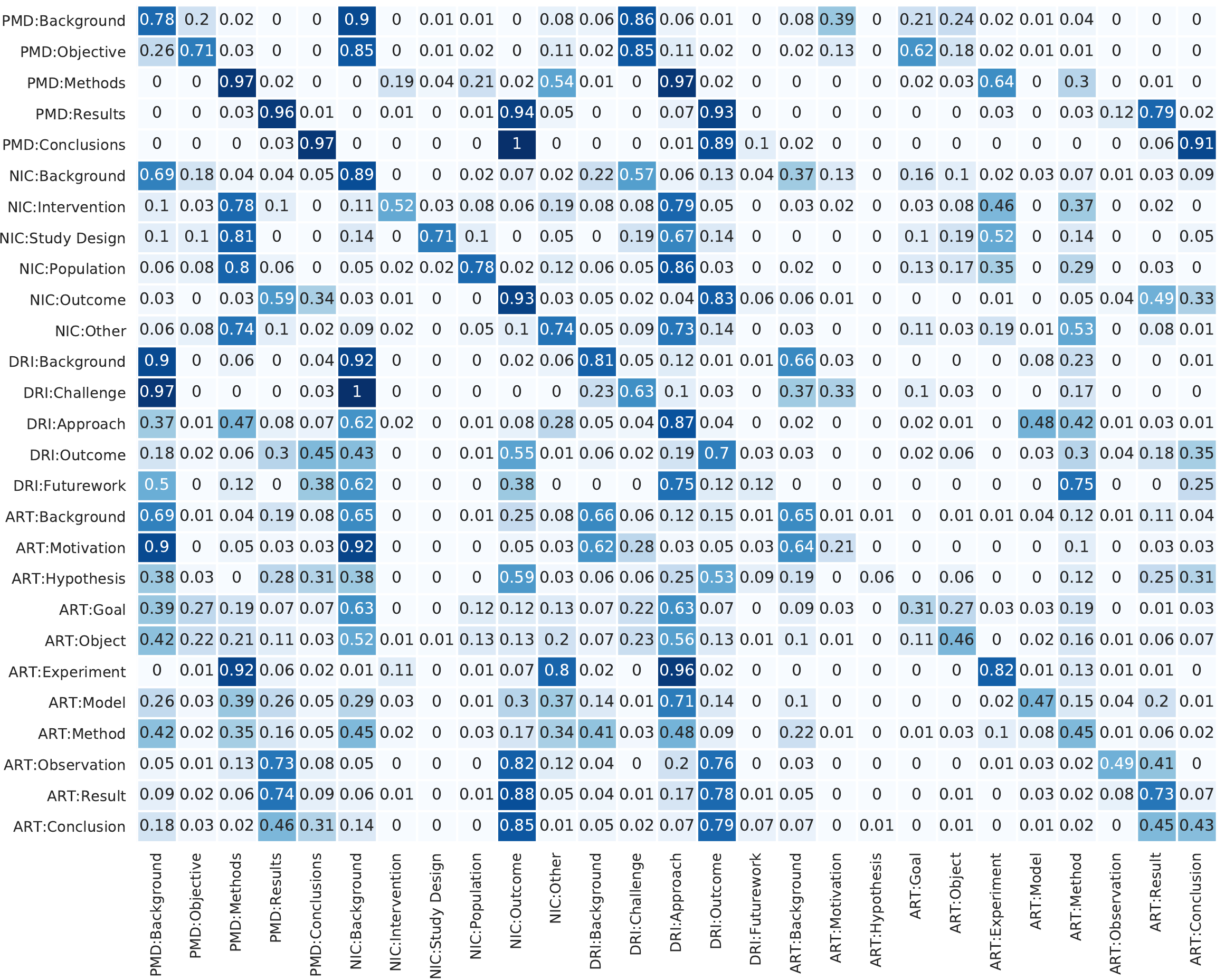}}
    \caption{Each row represents a semantic vector as described in Section~\ref{sec:sem_relatedness} for a class computed with \emph{MULT ALL} classifier.}
    \label{fig:pred_heatmap}
\end{figure*}

\paragraph{Effect of Dataset Size:}
The NIC and DRI models benefit more from multi-task learning than PMD and ART.
However, PMD and ART are bigger datasets than NIC and DRI.
The ART dataset has also more fine-grained classes than the other datasets. This raises the following question: 
\emph{How would the models for PMD and ART benefit from multi-task learning if they were trained on smaller datasets?} 

To answer this question, we created smaller variants of PMD and ART (i.e.\ $\mu$PMD and $\mu$ART) with a comparable size with NIC and DRI.
The training data was truncated to $\frac{1}{20}$ for $\mu$PMD and $\frac{1}{3}$ for $\mu$ART while keeping the original size of the validation and test sets.
As shown in Table~\ref{table:small_results}, all models benefit from the other datasets, whereas MULT GRP again performs best.
\emph{The results indicate that models for small datasets benefit from multi-task learning independent of the difference in the granularity of the classes (RQ1).}

\paragraph{Effect for each Class:}
Figure~\ref{fig:f1_per_label} shows the F1 scores per class for the investigated approaches.
Classes, which are intuitively highly semantically related (*:Background, *:Results, *:Outcome), and classes with few examples (DRI:FutureWork, DRI:Challenge, ART:Hypothesis, NIC:Study Design) tend to benefit significantly from multi-task learning.
The classes ART:Model, ART:\-Observation, and ART:Result have worse results than SciBERT-HSLN when using MULT ALL, but MULT GRP yields better results. 
This can be attributed to sharing the \emph{context enrichment layers} only between datasets with the same text type.
\emph{The analysis suggests that especially semantically related classes and classes with few examples benefit from multi-task learning (RQ1).}

\subsection{Semantic Relatedness of Classes across Annotation Schemes}
\label{sec:analysis_semantic_relatedness}
In this section, we first evaluate our proposed approach for the semi-automatical identification of semantically related classes in the datasets PMD, NIC, DRI, and ART. Based on the analysis, we identify six clusters of semantically related classes.
Then, we present a new dataset that is compiled from the investigated datasets and is based on the identified clusters.
As a possible down-stream application, this multi-domain dataset with a generic set of classes could help to structure research papers in a domain-independent manner, supporting, for instance, the development of academic search engines.

\begin{figure}[t!]
    \center{\includegraphics[width=1.0\linewidth]
        {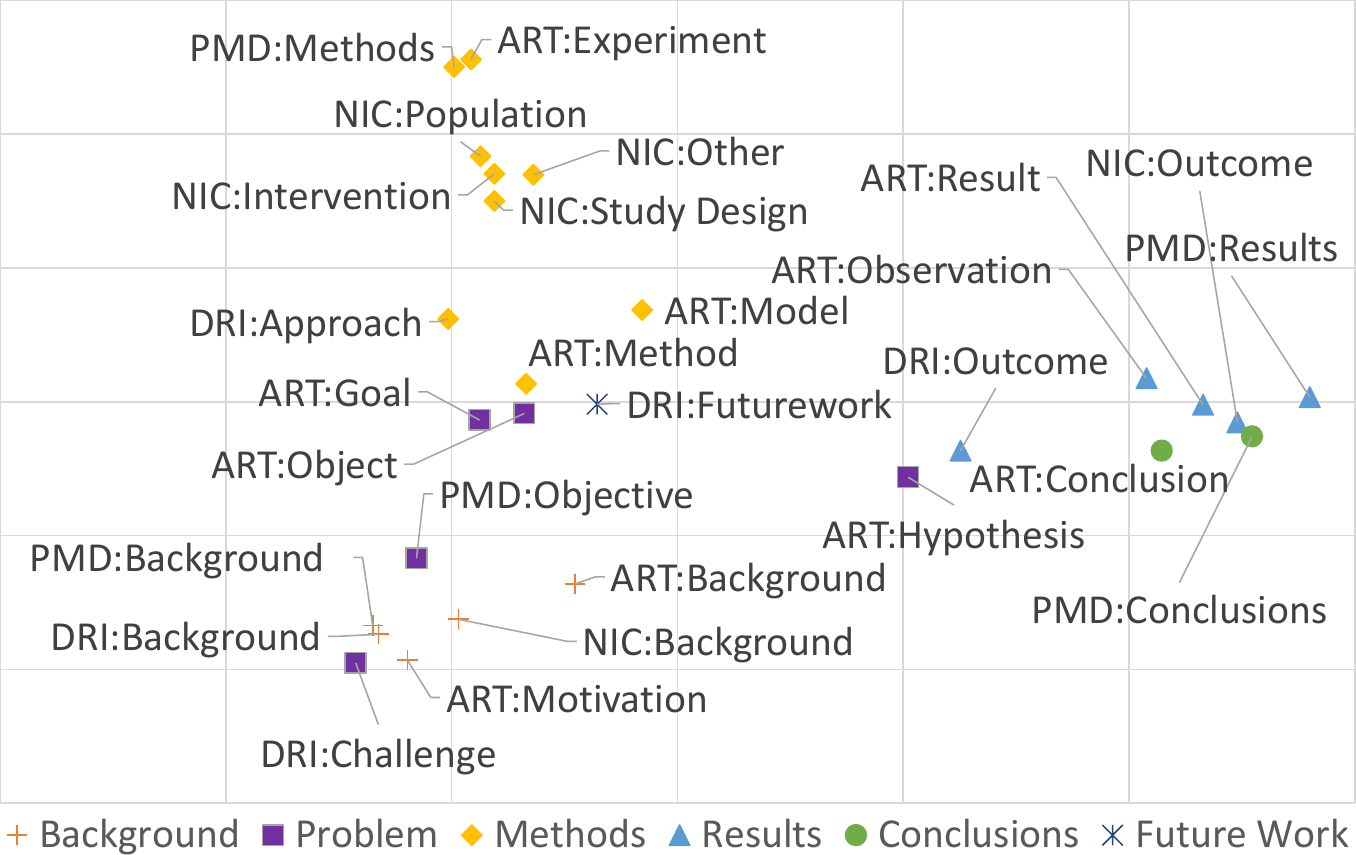}}
    \caption{Semantic vectors of labels computed with MULT ALL, and projected to 2D space via PCA. The semantic vectors are assigned to generic clusters of semantically related labels.}
    \label{fig:label_correlations}
\end{figure}

\paragraph{Analysis of Semantic Relatedness of Classes:}
Based on the annotation guidelines of the investigated datasets PMD~\citep{Dernoncourt2017PubMed2R}, NIC~\citep{Kim2011AutomaticCO}, DRI~\citep{Fisas2015OnTD}, and ART~\citep{Liakata2010CorporaFT}, we identified six clusters of semantically related classes, which are depicted in Figure~\ref{fig:label_correlations}. 
The identification process of the clusters followed the intuition, that most research papers independent of the scientific domain (1) investigate a research problem (\emph{Problem}), (2) provide background information for the problem (\emph{Background}), (3) apply or propose certain methods (\emph{Methods}), (4) yield results (\emph{Results}), (5) conclude the work (\emph{Conclusions}), and (6) outline future work (\emph{Future Work}).

\begin{table}[t!]
\center
\footnotesize
\caption{Silhouette scores per cluster and overall computed for the semantic vectors of SciBERT-HSLN, MULT GRP and MULT ALL classifiers.}
\label{tab:clusters_silhoutte}
\begin{tabular}{l|r|r|r}
                & SciBERT-HSLN & MULT GRP  & MULT ALL \\ \hline
Background      & 0.45         & 0.18     & \textbf{0.48}     \\
Problem         & -0.27        & \textbf{-0.04}    & -0.29    \\
Methods         & 0.19         & -0.03    & \textbf{0.31}     \\
Results         & -0.38        & 0.01     & \textbf{0.32}     \\ 
Conclusions     & \textbf{0.92}         & -0.49    & 0.02     \\
Future Work     & 0.00         & 0.00     & 0.00     \\
\hline
Overall  & 0.10         & -0.02    & \textbf{0.20}
\end{tabular}
\end{table}

Figure~\ref{fig:label_correlations} shows the semantic vectors for each label, which were computed with MULT ALL and projected to 2D space using principal component analysis~\cite{Jolliffe2011PrincipalCA}.
Each label is assigned to one of the generic clusters for semantically related labels.
Except \emph{Problem}, all clusters for semantically related classes are well identifiable in Figure~\ref{fig:label_correlations}.
The semantic vector for ART:Hypothesis is an outlier in the \emph{Problem} cluster because ART:Hypothesis is confused mostly with ART:Conclusion and ART:Result (see Figure~\ref{fig:pred_heatmap}) and has also a very low F1 score (see Figure~\ref{fig:f1_per_label}).
Table~\ref{tab:clusters_silhoutte} shows the Silhoutte scores~\cite{Rousseeuw1987} for each cluster. 
A positive score indicates that objects lie well within the cluster, and a negative score that the objects are merely somewhere in between clusters.
As a distance metric, we use $\text{semantic\_relatedness}$ as defined in Equation~\ref{eq:semantic_relatedness}.
The silhouette coefficients show that MULT ALL forms better clusters than SciBERT-HSLN and MULT GRP, although MULT GRP performs best.
We hypothesise that MULT ALL can better capture the semantic relatedness of labels than the other approaches since it is enforced to learn a generic feature extractor across multiple datasets.
\emph{The multi-task learning approach sharing all possible layers is able to recognise semantically related classes (RQ5).}

\paragraph{Domain-Independent Sentence Classification:}
Based on the identified clusters, we compile a new dataset \emph{G-PNDA} from the investigated datasets PMD, NIC, DRI, and ART.
The labels of the datasets are collapsed according to the clusters in Figure~\ref{fig:label_correlations}.
Table~\ref{table:generic_datasets} summarises the characteristics of the compiled dataset.
To prevent a bias towards bigger datasets, we truncate PMD to $\frac{1}{20}$ and ART to $\frac{1}{3}$ of their original size. 

Table~\ref{tab:generic_results} depicts our experimental results for the generic dataset \emph{G-PNDA}.
We train a model for each dataset part, and the multi-task learning models MULT ALL and MULT GRP. 
Since we have common sentence classes now, we train also models that share the output layers between the dataset parts, referred to as MULT ALL SHO and MULT GRP SHO (see Section~\ref{sec:transfer_methods}). 
For training and evaluation, we split each dataset into train/validation/test sets with the portions 70/10/20, average the results over three random restarts and use the same hyperparameters as before (see Section~\ref{sec:implementation}).

Table~\ref{tab:generic_results} shows that the proposed MULT GRP model outperforms all other settings.
Surprisingly, sharing the output layer impairs the performance in all settings. 
We can attribute this to the fact that the output layer learns different transition distributions between the classes. 
\emph{Thus, in a domain-independent setting a separate output layer per dataset part helps the model to capture the individual rhetorical structure present in the domains (RQ3, RQ6).}

\begin{table}[tb]
\footnotesize
\centering
\caption{Characteristics of the domain-independent dataset \emph{G-PNDA} that was compiled from the original datasets \emph{PMD}, \emph{NIC}, \emph{DRI}, and \emph{ART}.}
\begin{tabular}{l|r|r|r|r}
             & \textbf{G-PMD}   & \textbf{G-NIC}       & \textbf{G-DRI} & \textbf{G-ART}       \\
\hline
Text Type &  Abstract   & Abstract & Full paper & Full paper           \\
\hline
\# Papers &  1.000   & 1.000 & 40 & 67           \\
\# Sentences &  11.738    & 9.771  & 8.777 & 9.528           \\
$\varnothing$ \# Sentences  & 11  & 10   & 219  &  142          \\
\hline
Background          & 1.220 & 2.548 & 1.760     & 1.657  \\
Problem             & 953   & 0     & 449   & 529  \\
Methods             & 3.927 & 2.700 &  5.038     & 2.752  \\
Results              & 3.760 & 4.523 & 1.394 & 3.672  \\
Conclusions          & 1.878 & 0     & 0     & 918  \\
Future Work         & 0     & 0     & 136   & 0  \\
\hline
\end{tabular}
\label{table:generic_datasets}
\end{table}

\begin{table}[tb]
\footnotesize
\caption{Experimental results in terms of F1 scores for our proposed approaches for the generic dataset \emph{G-PNDA}:
baseline model SciBERT-HSLN with one separate model per dataset and the multi-task learning models MULT ALL SHO, MULT ALL, MULT GRP SHO, and MULT GRP.
\textbf{Bold} depicts whether the approach improves the baseline, \underline{underline} the best overall result.}        
\label{tab:generic_results}
\begin{tabular}{l|r|r|r|r|r}
                                 & G-PMD & G-NIC & G-DRI & G-ART & $\varnothing$ \\ 
\hline
SciBERT-HSLN                         & 90.1 & 89.3 & 81.7 & 70.8 & 83.0 \\
(one model per dataset)          &      &       &        &       & \\ 
\hline \hline
MULT ALL SHO                         & 89.8 & 89.1 & \textbf{83.5} & 67.1 & 82.4 \\
(shared output layer)               &      &       &        &     & \\
\hline
MULT ALL                       & \textbf{90.5} & \textbf{89.8}  & \textbf{84.9} & 70.5 & \textbf{83.9} \\
(separate output layer)         &      &       &        &  & \\
\hline \hline
MULT GRP SHO                        & 90.0 & \underline{\textbf{89.9}} & \textbf{86.1} & 70.4 & \textbf{84.1} \\
(shared output layer)          &      &       &        &   & \\
\hline
MULT GRP                        & \underline{\textbf{90.6}} & \textbf{89.7} & \underline{\textbf{87.2}} & \underline{\textbf{71.0}} & \underline{\textbf{84.6}} \\
(separate output layer)         &      &       &        &  &
\end{tabular}
\end{table}

\section{Conclusions}

In this paper, we have presented a unified deep learning architecture for sequential sentence classification.
The unified approach can be applied to datasets that contain abstracts as well as full articles.
For datasets of full papers, the unified approach significantly outperforms the state of the art without any feature engineering.

Furthermore, we have tailored two common transfer learning approaches to sequential sentence classification and compared their performance. 
We found that training a multi-task model with multiple datasets works better than sequential transfer learning. 
Our comprehensive experimental evaluation with four different datasets offers useful insights under which conditions transferring or sharing of specific layers is beneficial or not. 
In particular, it is always beneficial to share the sentence encoding layer between datasets from different domains.
However, it is most effective to share the context enrichment layer, which encodes the context of neighbouring sentences, only between datasets with the same text type (abstracts vs. full papers).
This can be attributed to different rhetorical structures in abstracts and full papers. 
Our tailored multi-task learning approach makes use of multiple datasets and yields new state-of-the-art results for two full paper datasets.
In particular, models for tasks with small datasets and classes with few labelled examples benefit significantly from models of other tasks.

Our study suggests that the classes of the different dataset annotation schemes are semantically related, even though the datasets come from different domains and have different text types (e.g. abstract or full papers). This semantic relatedness is an important prerequisite for transfer learning in NLP tasks~\citep{Mou2016HowTA,Pan10,Ruder2019Neural}, 

Finally, we proposed an approach to semi-automatically identify semantically related classes from different datasets to support manual comparison and inspection of different annotation schemes across domains. We demonstrated the usefulness of the approach with an analysis of four annotation schemes. This approach can support the investigation of annotation schemes across disciplines without re-annotating datasets.
From the analysis, we derived a domain-independent consolidated annotation scheme and compiled a domain-independent dataset. This allows for the classification of sentences in research papers with generic classes across disciplines, which can support, for instance, academic search engines.

In future work, we plan to integrate other tasks (e.g. scientific concept extraction) into the multi-task learning approach to exploit further datasets.
Furthermore, we intend to evaluate the domain-independent sentence classifier in an information retrieval scenario.

\bibliographystyle{ACM-Reference-Format}
\bibliography{references}

\end{document}